\documentclass[11pt]{article}

\PassOptionsToPackage{table}{xcolor}
\usepackage[final]{acl}

\usepackage{times}
\usepackage{latexsym}

\usepackage[T1]{fontenc}

\usepackage[utf8]{inputenc}

\usepackage{microtype}

\usepackage{inconsolata}

\usepackage{graphicx}

\usepackage{url}
\usepackage{booktabs}
\usepackage{subcaption}
\usepackage{tabularx}
\usepackage{amsmath,amsthm}
\usepackage{amsfonts}
\usepackage{bm}
\usepackage{nicefrac}
\usepackage{algorithm}
\usepackage{algpseudocode}
\usepackage{threeparttable}
\usepackage{multirow}

\makeatletter
\newcommand{\algwrap}[1]{\parbox[t]{\dimexpr\linewidth-\ALG@tlm-3em\relax}{\raggedright\strut #1\strut}}
\newcommand{\alghighlightwrap}[2]{\begingroup\setlength{\fboxsep}{1pt}\colorbox{#1}{\parbox[t]{\dimexpr\linewidth-\ALG@tlm-3em-2\fboxsep\relax}{\raggedright\strut #2\strut}}\endgroup}
\makeatother
\newcommand{\StateW}[1]{\State \algwrap{#1}}

\newcommand{\alginlinehighlight}[2]{\begingroup\setlength{\fboxsep}{1pt}\colorbox{#1}{#2}\endgroup}
\newcommand{\StateYellow}[1]{\State \alghighlightwrap{yellow!30}{#1}}
\newcommand{\StateBlue}[1]{\State \alghighlightwrap{blue!12}{#1}}
\newcommand{\IfYellow}[1]{\If{\alginlinehighlight{yellow!30}{#1}}}
\newcommand{\IfBlue}[1]{\If{\alginlinehighlight{blue!12}{#1}}}
\newcommand{\ElseYellow}{\algrenewcommand\algorithmicelse{\alginlinehighlight{yellow!30}{\textbf{else}}}\Else\algrenewcommand\algorithmicelse{\textbf{else}}}
\newcommand{\ElseBlue}{\algrenewcommand\algorithmicelse{\alginlinehighlight{blue!12}{\textbf{else}}}\Else\algrenewcommand\algorithmicelse{\textbf{else}}}
\DeclareRobustCommand{\algcaptionyellow}[1]{\begingroup\setlength{\fboxsep}{1pt}\colorbox{yellow!30}{#1}\endgroup}
\DeclareRobustCommand{\algcaptionblue}[1]{\begingroup\setlength{\fboxsep}{1pt}\colorbox{blue!12}{#1}\endgroup}

\theoremstyle{definition}

\theoremstyle{plain}

\title{FedSubMuon: Communication-Efficient Federated \\
  LLM Fine-Tuning via Structured Subspace Muon}

\author{
\textbf{Shaolong Chen\textsuperscript{1,2,3}},
\textbf{Youming Tao\textsuperscript{4}},
\textbf{Shuzhen Chen\textsuperscript{5}},
\textbf{Falko Dressler\textsuperscript{4}},
\textbf{Qingqing Ye\textsuperscript{6}},
\textbf{Di Wang\textsuperscript{1,2}\thanks{Corresponding author.}}\\
\textsuperscript{1}King Abdullah University of Science and Technology (KAUST), \\
\textsuperscript{2}Provable Responsible AI and Data Analytics (PRADA) Lab,\\
\textsuperscript{3}Imperial College London,
\textsuperscript{4}Technische Universit\"{a}t Berlin (TU Berlin),\\
\textsuperscript{5}Ocean University of China,
\textsuperscript{6}The Hong Kong Polytechnic University
}

\begin{document}

\maketitle

\begin{abstract}
Federated fine-tuning adapts large language models (LLMs) to decentralized
client data, but its scalability in cross-device training is often limited
by the high communication cost. Muon is an
optimizer that improves optimization performance by orthogonalizing momentum for matrix-valued parameters.
Existing federated Muon methods demonstrate the benefit of matrix-aware
optimization in federated learning, but still require transmitting
full layer-size updates
and optimizer state. A natural way to reduce communication is to directly apply Muon to LoRA
factors, but this changes the optimized object and weakens Muon's
matrix-aware update geometry.
We propose \textsc{FedSubMuon}, a communication-efficient federated
Muon fine-tuning method that optimizes compact coefficient matrices
within shared structured subspaces. This design keeps Muon on a single matrix-valued trainable object,
while reducing the client upload to compact
coefficient matrices. We further introduce \textsc{FedSubMuon-GT},
an accuracy-oriented extension that uses projected gradients to adapt tracked subspace bases toward
task-relevant gradient directions. Experiments on instruction tuning and mathematical reasoning
show that \textsc{FedSubMuon-GT} achieves the best overall accuracy on four of
five dataset--model pairs, while \textsc{FedSubMuon} performs best under all matched
communication budgets. On Dolly-15K, the closest communication baseline
requires \(5.5\times\) and \(1.4\times\) more total communication on Llama-1B
and Qwen-4B, respectively.
\end{abstract}

\section{Introduction}

Federated fine-tuning adapts large language models (LLMs) on decentralized
client data without transferring raw examples to a central server. In
cross-device settings, however, this adaptation is often constrained by communication cost:
in each communication round, the server sends trainable parameters
to selected clients, and clients return local updates for
aggregation \citep{zhang2024fedit,liu2025ecolora}. As a result, the size of the
communicated update becomes a central design constraint for federated LLM
fine-tuning.

\begin{table}[!t]
\centering
\small
\setlength{\tabcolsep}{8pt}
\renewcommand{\arraystretch}{1.05}
\caption{Optimizer control for LoRA-factor federated baselines on Natural
Instructions with Llama 1B. Each cell reports mean ROUGE-L\% over two runs.}
\label{tab:intro_lora_muon_control}
\vspace{-0.3em}
\begin{tabular}{lccc}
\toprule
Method & AdamW & Muon & SGD \\
\midrule
FedIT
& $30.96$
& $30.81$
& $27.94$ \\
FLoRA
& $31.14$
& $30.99$
& $27.85$ \\
\bottomrule
\end{tabular}
\vspace{-2em}
\end{table}

Muon is a matrix-aware optimizer that improves optimization performance by orthogonalizing momentum before updating
hidden-layer weights \citep{jordan2024muon}. Existing work has begun to adapt
Muon to federated learning, including FedMuon variants for
matrix-orthogonalized federated optimization
\citep{liu2025fedmuon,takezawa2025fedmuon,zhang2025provablefedmuon,wang2026canzona}.
This line of work is relevant because exploiting the matrix structure of model
updates has been shown to improve optimization performance across a range of
federated settings. However, existing
FedMuon work does not remove the communication bottleneck for
fine-tuning: applying Muon to full weight matrices requires transmitting
full layer-size parameter updates.

A natural alternative is to apply Muon within parameter-efficient federated fine-tuning methods like LoRA, but this
changes the object being optimized.
LoRA represents an update through two coupled low-rank factors, and prior work
shows that applying Muon to these factors is not equivalent to a
parametrization-independent update of the underlying matrix
\citep{bogachev2026riemannion}. Consistent with this issue,
Table~\ref{tab:intro_lora_muon_control} shows that replacing the optimizer in
federated LoRA baselines with Muon does not improve performance.
These observations motivate a federated Muon fine-tuning method that
preserves Muon's matrix update geometry while communicating only a compact
trainable object.

We propose \textsc{FedSubMuon}, a communication-efficient framework for federated
Muon fine-tuning with a structured subspace. \textsc{FedSubMuon} reduces communication by
transmitting a compact coefficient matrix in a shared matrix subspace.
Clients optimize the coefficient matrix with Muon, and upload it for aggregation.
At refresh rounds, the server commits the aggregated subspace update into a
global accumulator and samples a new coefficient matrix.
Then we introduce
\textsc{FedSubMuon-GT}, which uses projected gradients at refresh rounds
to update tracked subspace bases. While \textsc{FedSubMuon} emphasizes communication efficiency,
\textsc{FedSubMuon-GT} adds basis-tracking communication to improve accuracy.

Experiments on instruction tuning and mathematical reasoning with Llama and
Qwen backbones show this trade-off. \textsc{FedSubMuon-GT} achieves the strongest overall
accuracy, ranking first on four of five dataset--model pairs. Under matched communication budgets, \textsc{FedSubMuon} performs best on
both evaluated backbones, showing that simply lowering the rank of LoRA
baselines is not a reliable substitute for the proposed
subspace coefficient parameterization.

Our contributions are:
\vspace{-0.7em}
\begin{itemize}
    \setlength{\topsep}{0.15em}
    \setlength{\partopsep}{0pt}
    \setlength{\itemsep}{-0.3em}
    \setlength{\parsep}{0pt}
    \item We propose \textsc{FedSubMuon}, which adapts federated Muon to
    communication-efficient fine-tuning by transmitting compact
    coefficient matrices rather than full weight matrices.
    \item We develop \textsc{FedSubMuon-GT}, a performance-oriented extension
    that uses projected gradients to adapt tracked subspace bases to task-relevant
    gradient directions.
    \item We evaluate the proposed method on instruction tuning and
    mathematical reasoning across 1-8B LLMs, demonstrating improved
    accuracy--communication trade-offs under standard and matched-budget
    settings.
\end{itemize}

\section{Preliminaries}

\paragraph{Federated fine-tuning.}
We consider cross-device federated fine-tuning of a frozen pretrained backbone \(W^0\) over \(N\) clients,
where private data remain on clients~\citep{mcmahan2017fedavg}. Client \(i\) owns a local objective \(f_i\), and the global objective is
\[
\min_{\Delta W} f(\Delta W)
:=
\sum_{i=1}^{N} p_i f_i(W^0+\Delta W),
\]
where \(\mathcal L\) denotes the index set of selected weight matrices,
\(\Delta W=\{\Delta W_\ell\}_{\ell\in\mathcal L}\) is the update on those
matrices, and \(p_i\) is the aggregation weight of client \(i\). 
At round \(t\), we write \(W^t:=W^0+\Delta W^t\) for the current model.
A prevalent federated fine-tuning strategy adapts selected projection
matrices with low-rank matrices~\citep{hu2022lora}; selected clients train and
upload adapter parameters for server
aggregation~\citep{zhang2023fedpetuning,zhang2024fedit}.

\paragraph{Muon Optimizer.}
Muon is a matrix-aware optimizer that updates hidden-layer weight matrices by
orthogonalizing the momentum~\citep{jordan2024muon}. Consider the full-matrix
update for a layer \(\ell\).
With global gradient \(G_\ell^t=\nabla_{W_\ell}f(W^t)\) and matrix momentum \(M_\ell^t\), full Muon would use
\[
\begin{aligned}
M_\ell^{t+1}
&=
(1-\beta)M_\ell^t+\beta G_\ell^t,
\\
W_\ell^{t+1}
&=
W_\ell^t-\eta\,\operatorname{Muon}(M_\ell^{t+1}),
\end{aligned}
\]
where \(\operatorname{Muon}(A)=\widetilde U\widetilde V^\top\) for the compact SVD \(A=\widetilde U\Sigma\widetilde V^\top\); Newton–Schulz (NS) provides an efficient approximation to this direction practically.

\paragraph{Random subspace.}
Random subspace optimization restricts training to a low-dimensional
coordinate inside the full parameter
space~\citep{li2018intrinsic,aghajanyan2021intrinsic}. For a layer \(\ell\), a
generic vectorized random subspace is
\[
\operatorname{vec}(\Delta W_\ell)=P_\ell b_\ell,
\]
where \(P_\ell\in\mathbb R^{d_{\mathrm{out}}d_{\mathrm{in}}\times k}\) is a random projection matrix
and \(b_\ell\in\mathbb R^k\) is the trainable subspace coordinate. The projection
matrix can be resampled
periodically~\citep{yang2026fedkrso,rajabi2025subtrack}; switching \(P_\ell\)
changes the low-dimensional subspace used by subsequent updates.

\section{Method}
\label{sec:method}
\label{sec:fedsubmuon}
\label{sec:fedsubmuon_gt}

\subsection{Overview}
\label{sec:method_overview}

\textsc{FedSubMuon} is a structured subspace method for federated Muon
fine-tuning. The backbone remains frozen, while the server maintains a global
update accumulator over the selected weight matrices. Clients train compact
subspace coefficient matrices with fixed bases generated from shared
random seeds. This parameterization applies Muon to a matrix-structured
coefficient update while reducing the per-round trainable parameter count and
communication payload.

Figure~\ref{fig:core-update} illustrates the key workflow of \textsc{FedSubMuon}. The
server samples clients, sends the current random seed, active subspace coefficient matrix
and any missing commit records to clients. A selected client first replays those commits to synchronize the
global update accumulator to its local accumulator. It then regenerates the shared
bases from the random seed, runs local Muon updates on the compact subspace coefficient matrix, and
uploads the updated matrix for server aggregation.

Refresh rounds separate accumulated model updates from later subspace updates.
At the refresh round, the server commits the full model update into the global update
accumulator, appends the aggregated coefficient matrix and corresponding seeds to the commit log,
resets the coefficient matrix, and samples fresh random seeds for a new
basis. If the current communication round is not a refresh round, the
server keeps the global update accumulator and random seed unchanged and sets the
aggregated value as the active subspace coefficient matrix for the next communication
round.

Appendix~\ref{app:unified_algorithm} provides the full procedural descriptions
of both \textsc{FedSubMuon} and \textsc{FedSubMuon-GT}, corresponding to
Algorithms~\ref{alg:fedsubmuon} and~\ref{alg:fedsubmuon_gt}.

\subsection{Client Synchronization and Subspace Coefficient Update}
\label{sec:compact_core_update}

\begin{figure}[t]
    \centering
    \includegraphics[width=\columnwidth]{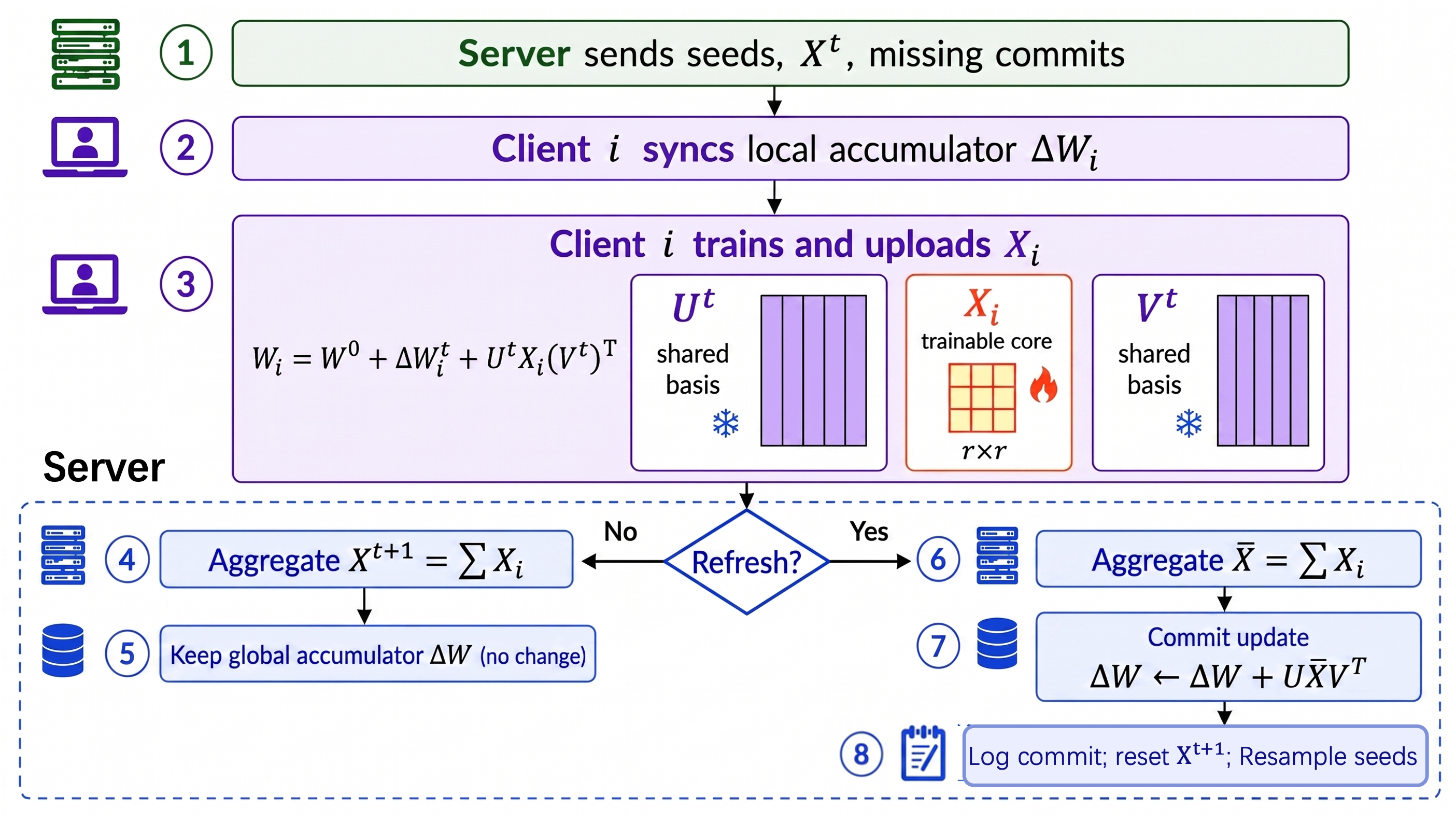}
    \caption{\textsc{FedSubMuon} Workflow.}
    \label{fig:core-update}
    \vspace{-1em}
\end{figure}

We first describe one communication round before separating the refresh branch.
A selected client first synchronizes its local accumulator by
replaying missing commits; since those records are created at refresh
rounds, we give the replay rule in Section~\ref{sec:commit_reset}. After this
synchronization, all selected clients start local training from the same
subspace coefficient matrices \(X^t\) and use the same bases.

For a selected layer \(\ell\), the update used in the current round is separated
into two parts: the synchronized local accumulator \(\Delta W_{i,\ell}\)
and the active subspace coefficient matrix \(X_{i,\ell}\). The former stores
committed matrix updates, while the latter is the only trainable object during
the current round. In \textsc{FedSubMuon}, the current bases are generated from
the random seed \(s^t\). For each selected layer, the shared seed
deterministically generates Gaussian matrices
\(\Omega_{U,\ell}^t\in\mathbb R^{d_{\mathrm{out}}\times r}\) and
\(\Omega_{V,\ell}^t\in\mathbb R^{d_{\mathrm{in}}\times r}\) with i.i.d.
\(\mathcal N(0,1)\) entries, and
\[
U_\ell^t=\operatorname{qf}(\Omega_{U,\ell}^t),
\qquad
V_\ell^t=\operatorname{qf}(\Omega_{V,\ell}^t),
\]
where \(\operatorname{qf}(\cdot)\) returns the thin-QR orthonormal factor with a
fixed deterministic sign convention.

With the current bases fixed, client \(i\) optimizes only the subspace
coefficient matrix \(X_{i,\ell}\in\mathbb R^{r\times r}\). The effective local
layer used for training is
\begin{equation}
\label{eq:core_lift}
W_{i,\ell}(X_{i,\ell})
=
W_\ell^0+\Delta W_{i,\ell}
+
U_\ell^t X_{i,\ell}(V_\ell^t)^\top .
\end{equation}
Eq.~\eqref{eq:core_lift} is the local training model: the frozen backbone \(W_\ell^0\), the
fixed synchronized accumulator \(\Delta W_{i,\ell}\) and fixed subspace bases \((U_\ell^t, V_\ell^t)\), only
\(X_{i,\ell}\) is updated. At the start of local optimization, the client initializes
\begin{equation}
\label{eq:core_init}
X_{i,\ell}^{t,0}=X_\ell^t,\qquad M_{i,\ell}^{t,0}=0 .
\end{equation}
For local step \(e\), it computes a stochastic gradient with respect to the
coefficient matrix,
\begin{equation}
\label{eq:core_gradient}
\widehat G_{i,\ell}^{t,e}
=
\nabla_{X_{i,\ell}}
f_i(W_i^t(X_i);\xi_i^{t,e})
\big|_{X_i=X_i^{t,e}} .
\end{equation}
By the chain rule, this is the current layer gradient projected onto the
current left and right bases: when the full layer gradient is written as
\(\nabla_{W_{i,\ell}} f_i\), the corresponding coefficient gradient has the
form
$(U_\ell^t)^\top
(\nabla_{W_{i,\ell}} f_i)
V_\ell^t$.
Thus Eq.~\eqref{eq:core_gradient} updates the \(r\times r\) coefficient matrix,
not a full layer-size matrix.

The client then updates the local momentum and applies Muon to this small
matrix:
\begin{equation}
\label{eq:core_muon_update}
\begin{aligned}
M_{i,\ell}^{t,e+1}
&=(1-\beta)M_{i,\ell}^{t,e}
+\beta\widehat G_{i,\ell}^{t,e},\\
X_{i,\ell}^{t,e+1}
&=X_{i,\ell}^{t,e}
-\eta\,\operatorname{Muon}(M_{i,\ell}^{t,e+1}).
\end{aligned}
\end{equation}
This local update corresponds to step 3 in
Figure~\ref{fig:core-update}. After \(E\) local steps, the client sets
\(X_{i,\ell}^{t+1}=X_{i,\ell}^{t,E}\). The momentum is reinitialized whenever
the client is selected for a new round, so it remains local \(r\times r\)
optimizer state.

The client uploads \(\{X_{i,\ell}^{t+1}\}_{\ell\in\mathcal L}\) and the server
aggregates the uploaded coefficient matrices with the normalized round weights,
$
\bar X_\ell^{t+1}
=
\sum_{i\in\mathcal S_t}a_i^t X_{i,\ell}^{t+1}, \ell\in\mathcal L .
$
Since all selected clients start from the same active coefficient matrix
\(X_\ell^t\), this aggregation averages compact coefficient updates inside the
same current subspace.

\subsection{Commit-and-Reset Refresh}
\label{sec:commit_reset}

After aggregation, the server checks the refresh interval \(\tau\) to determine
whether this round performs a refresh, as shown in
Figure~\ref{fig:core-update}. If the round is not a refresh
round, the server keeps the global accumulator and the current bases
unchanged, set:
\[
\Delta W_\ell^{t+1}\leftarrow \Delta W_\ell^t,
\qquad
X_\ell^{t+1}\leftarrow \bar X_\ell^{t+1}.
\]
In \textsc{FedSubMuon}, the same random seed is also kept for the next round.
Thus non-refresh rounds simply continue optimizing the active coefficient
matrix in the current subspace.

At a refresh round, the server commits the active coefficient update to the
global accumulator and resets the active coefficient matrix:
\begin{equation}
\label{eq:commit_reset_update}
\Delta W_\ell^{t+1}
\leftarrow
\Delta W_\ell^t
+
U_\ell^t \bar X_\ell^{t+1}(V_\ell^t)^\top,
X_\ell^{t+1}\leftarrow 0 .
\end{equation}
Eq.~\eqref{eq:commit_reset_update} is step 7 in the
refresh branch of Figure~\ref{fig:core-update}. The commit uses the bases that were fixed during
the current round, because those bases define the meaning of
\(\bar X_\ell^{t+1}\). After the commit, \(\Delta W_\ell^{t+1}\) stores the
accumulated matrix update for layer \(\ell\), while the reset
\(X_\ell^{t+1}\leftarrow 0\) starts the next subspace update from zero.

This separation is important because the global update accumulator after
multiple refresh rounds is a sum of matrix updates produced under different
bases. A newly sampled random basis generally cannot represent that accumulated
update exactly as a fresh coefficient matrix. Therefore, \textsc{FedSubMuon}
keeps the committed update in the global update accumulator
\(\Delta W_\ell^t\), and uses the next basis only to parameterize subsequent
coefficient updates. After a refresh, \textsc{FedSubMuon} samples fresh random
seeds \(s^{t+1}\), and clients in the next round regenerate the shared bases
using the Gaussian--QR construction in Section~\ref{sec:compact_core_update}.

Refreshes are synchronized lazily to handle partial participation. In
\textsc{FedSubMuon}, the server appends each refresh to the compact log
\(\mathcal H\) as
\[
\mathcal C_K
=
\bigl(
s^{[K]}=s^t,\,
\{X_\ell^{[K]}=\bar X_\ell^{t+1}\}_{\ell\in\mathcal L}
\bigr),
\]
and increments the commit counter \(K\). The bracketed superscript denotes a
commit index, not a communication round. If client \(i\) was inactive during
some refresh rounds, it does not receive those commits immediately. When it is
selected again, the server sends the missing records
\(\{\mathcal C_k\}_{k=v_i}^{K-1}\). For each missing record, the client
regenerates the committed bases from \(s^{[k]}\) and replays
\[
\Delta W_{i,\ell}
\leftarrow
\Delta W_{i,\ell}
+
U_\ell^{[k]}X_\ell^{[k]}(V_\ell^{[k]})^\top,
\qquad \ell\in\mathcal L .
\]
It then sets \(v_i\leftarrow K\). This synchronization is the step 2 in Figure~\ref{fig:core-update}
performed before Eq.~\eqref{eq:core_lift}.

\subsection{\textsc{FedSubMuon-GT}: Gradient-Tracked Subspace Refresh}
\label{sec:gt_refresh}

\textsc{FedSubMuon} uses random refresh to explore new search directions, but the refreshed bases
are task-agnostic: they do not use the gradients observed during training,
so the effective capture of task-relevant ambient directions can be weak.
\textsc{FedSubMuon-GT} improves this by keeping the same client-side coefficient update,
aggregation, and commit-and-reset rule as \textsc{FedSubMuon}, while replacing
random basis refresh with gradient-tracked basis refresh. At initialization, the bases are
generated from the random seed \(s^0\). But after the first refresh, the server sends tracked subspace bases
\((U_\ell^t,V_\ell^t)\) instead of sampling a fresh seed.
The goal is to make subspace coefficient updates better capture task-relevant
full-matrix directions by steering future bases using client gradients.

Updating tracked bases from a full layer gradient would require transmitting
full layer-size gradient information. To avoid this cost, selected clients send
projected gradients at refresh rounds. After completing the local coefficient
update, client \(i\) evaluates the updated local model \(W_i^t(X_i^{t+1})\) on
a small probe batch, uses the resulting layer gradient as the probe
gradient, forms
\begin{equation}
\label{eq:gt_gradient_summaries}
\begin{aligned}
H_{U,i,\ell}^{t}
&=
G_{i,\ell}^{\mathrm{probe},t}V_\ell^t,
&
H_{V,i,\ell}^{t}
&=
(G_{i,\ell}^{\mathrm{probe},t})^\top U_\ell^t ,
\end{aligned}
\end{equation}
where \(G_{i,\ell}^{\mathrm{probe},t}\) denotes the local gradient used for
basis tracking. In non-refresh rounds, clients upload only
\(\{X_{i,\ell}^{t+1}\}_{\ell\in\mathcal L}\). In refresh rounds, they upload
\(\{X_{i,\ell}^{t+1},H_{U,i,\ell}^t,H_{V,i,\ell}^t\}_{\ell\in\mathcal L}\).
The projected gradients keep the basis
update dependent on the current local gradient, while avoiding transmission of
a full gradient matrix.

The server aggregates these projected gradients with the same
weights \(a_i^t\) used to aggregate \(X\):
\begin{equation}
\label{eq:gt_summary_aggregate}
\begin{aligned}
\bar H_{U,\ell}^{t}
&=
\sum_{i\in\mathcal S_t}a_i^t H_{U,i,\ell}^{t},
&
\bar H_{V,\ell}^{t}
&=
\sum_{i\in\mathcal S_t}a_i^t H_{V,i,\ell}^{t}.
\end{aligned}
\end{equation}
By linearity, these summaries are equivalent to projecting the weighted probe
gradient
\(\bar G_\ell^t=\sum_{i\in\mathcal S_t}a_i^t
G_{i,\ell}^{\mathrm{probe},t}\) onto the current opposite-side bases:
$\bar H_{U,\ell}^{t}=\bar G_\ell^t V_\ell^t$ and 
$\bar H_{V,\ell}^{t}=(\bar G_\ell^t)^\top U_\ell^t$.

The basis update follows the principle of gradient subspace tracking. Given the
current right basis \(V_\ell^t\), the left basis should be adjusted to better
span the dominant directions of \(\bar G_\ell^t V_\ell^t\); the right basis is
handled analogously. This can be viewed as two small
least-squares subspace fitting problems:
\[
\begin{aligned}
\mathcal J_U(U)
&=
\min_{B_{U,\ell}} \|UB_{U,\ell}-\bar H_{U,\ell}^{t}\|_F^2,
\\
\mathcal J_V(V)
&=
\min_{B_{V,\ell}} \|VB_{V,\ell}-\bar H_{V,\ell}^{t}\|_F^2 .
\end{aligned}
\]
Since \(U_\ell^t\) and \(V_\ell^t\) are orthonormal, the least-squares
coordinates at the current bases are
\begin{subequations}
\label{eq:gt_basis_tracking}
\begin{equation}
\label{eq:gt_ls_coordinates}
B_{U,\ell}^{t}=(U_\ell^t)^\top \bar H_{U,\ell}^{t},
\qquad
B_{V,\ell}^{t}=(V_\ell^t)^\top \bar H_{V,\ell}^{t}.
\end{equation}
The corresponding residual components outside the current bases are
\((I-U_\ell^t(U_\ell^t)^\top)\bar H_{U,\ell}^{t}\) and
\((I-V_\ell^t(V_\ell^t)^\top)\bar H_{V,\ell}^{t}\). Following the
Grassmannian subspace-tracking view \citep{rajabi2025subtrack}, where a subspace is updated using the
residual of a least-squares gradient approximation rather than recomputed from
scratch, we use the following first-order tangent directions:
\begin{equation}
\label{eq:gt_residual_direction}
\begin{aligned}
Y_{U,\ell}^{t}
&=
(I-U_\ell^t(U_\ell^t)^\top)
\bar H_{U,\ell}^{t}(B_{U,\ell}^{t})^\top,\\
Y_{V,\ell}^{t}
&=
(I-V_\ell^t(V_\ell^t)^\top)
\bar H_{V,\ell}^{t}(B_{V,\ell}^{t})^\top .
\end{aligned}
\end{equation}
\end{subequations}
Here the projection matrices remove the components already represented by the
current bases, while multiplication by
\((B_{U,\ell}^{t})^\top\) and \((B_{V,\ell}^{t})^\top\) weights each residual
direction by its interaction with the currently represented coefficient-gradient
coordinates. Therefore, \(Y_{U,\ell}^{t}\) and \(Y_{V,\ell}^{t}\) do not restart
the subspace; instead, they rotate the existing bases toward gradient directions
that are useful but currently underrepresented.

Finally, we update the bases:
\begin{equation}
\label{eq:gt_basis_retraction}
\begin{aligned}
U_\ell^{t+1}
&=
\operatorname{qf}(U_\ell^t+\rho Y_{U,\ell}^{t}),\\
V_\ell^{t+1}
&=
\operatorname{qf}(V_\ell^t+\rho Y_{V,\ell}^{t}).
\end{aligned}
\end{equation}
The subspace learning rate \(\rho\) controls the size of the basis rotation, and
the QR retraction restores orthonormality. This gives a lightweight
first-order approximation to a Grassmannian subspace update, adapted to the
structured subspace. The update is
applied after committing the current coefficient update, so the basis refresh
tracks the gradient directions that are not captured by the already accumulated
subspace update.

At the same refresh round, \textsc{FedSubMuon-GT} appends a commit record with
the tracked bases used for the committed update:
$
\mathcal C_K
=
\bigl(
\{U_\ell^{[K]}=U_\ell^t,\,
V_\ell^{[K]}=V_\ell^t,\,
X_\ell^{[K]}=\bar X_\ell^{t+1}\}_{\ell\in\mathcal L}
\bigr).
$
Thus, inactive clients can later replay the exact committed update into their
local update accumulators. The difference from \textsc{FedSubMuon} is that
\textsc{FedSubMuon} stores the random seed used to regenerate the committed
bases, whereas \textsc{FedSubMuon-GT} stores the tracked bases themselves.

\subsection{Communication Analysis}
\label{sec:method_cost}

In each communication round of \textsc{FedSubMuon}, the server sends the current random seed \(s^t\), the
active subspace coefficient matrices \(X^t\), and any missing commit records.
For each selected layer, a standard round sends an \(r\times r\) coefficient
matrix to each selected client and receives the updated coefficient matrix back.
Therefore the per-client total communication for
the active coefficient matrix is \(O(r^2)\) downlink plus \(O(r^2)\)
uplink, independent of the full layer matrix size.
Note that \(r\) is the side length of a coefficient matrix decoupled from the
layer dimensions, whereas the LoRA rank indexes an adapter tied to
\(d_{\mathrm{out}}\) and \(d_{\mathrm{in}}\), so the two are not comparable on a
shared rank axis. With \(r\ll\min\{d_{\mathrm{out}},d_{\mathrm{in}}\}\), the
per-round payload stays at or below the LoRA baselines
(Appendix Table~\ref{tab:trainable_param_counts}).
This communication reduction does not imply that the committed update is stored in a factorized form:
the global update accumulator \(\Delta W_\ell^t\) is an ambient matrix for each selected layer,
and synchronized clients may materialize the corresponding local accumulator \(\Delta W_{i,\ell}\) during training.
At deployment time, the accumulated update can be merged into the frozen backbone,
so our method does not require an additional full-layer-size parameter store.

Refresh rounds add only compact synchronization state in \textsc{FedSubMuon}. The server
records
\(\mathcal C_K=(s^{[K]}=s^t,\{X_\ell^{[K]}=\bar X_\ell^{t+1}\}_{\ell\in\mathcal L})\)
in the commit log and samples fresh random seeds. Inactive clients receive and
replay missed commit records when they are selected again. \textsc{FedSubMuon-GT} follows the
same communication path, but after the first refresh it sends tracked
subspace bases \((U_\ell^t,V_\ell^t)\) in post-refresh communication rounds; its
commit records store these bases together with the
coefficient matrix. This tracked-basis downlink scales as
\(O((d_{\mathrm{out}}+d_{\mathrm{in}})r)\) whenever it is sent and can dominate
the total traffic. At refresh rounds, selected clients also upload projected
gradients \(H_{U,i,\ell}^t\) and \(H_{V,i,\ell}^t\), which add an amortized
\(O((d_{\mathrm{out}}+d_{\mathrm{in}})r/\tau)\) uplink term when refreshes occur
every \(\tau\) rounds.

\begin{table*}[!t]
\centering
\caption{Performance comparison across datasets (\%). Each cell reports mean $\pm$
standard deviation over four runs.
\textbf{Best} and \underline{second-best} values are highlighted.}
\label{tab:main_results_updated}
\small
\setlength{\tabcolsep}{4pt}
\renewcommand{\arraystretch}{0.98}
\setlength{\aboverulesep}{0.35ex}
\setlength{\belowrulesep}{0.35ex}

\newcommand{\best}[1]{\textbf{#1}}
\newcommand{\second}[1]{\underline{#1}}

\begin{tabular*}{\textwidth}{@{\extracolsep{\fill}}lccccc@{}}
\toprule
\multirow{2}{*}{Method}
& \multicolumn{1}{c}{Natural Instructions}
& \multicolumn{2}{c}{Dolly-15K}
& \multicolumn{1}{c}{GSM8K}
& \multicolumn{1}{c}{MATH} \\
\cmidrule(lr){2-2}\cmidrule(lr){3-4}\cmidrule(lr){5-5}\cmidrule(lr){6-6}
& Llama 1B & Llama 1B & Qwen 4B & Qwen 8B & Qwen 8B \\
\midrule

FedIT
& 30.14 $\pm$ 0.82
& 28.32 $\pm$ 0.41
& 36.33 $\pm$ 0.17
& 79.59 $\pm$ 0.33
& 45.24 $\pm$ 1.22 \\

FeDeRA
& 31.07 $\pm$ 0.43
& 29.63 $\pm$ 0.53
& 36.78 $\pm$ 0.37
& 80.29 $\pm$ 0.26
& 47.52 $\pm$ 0.65 \\

FLoRA
& \second{31.12 $\pm$ 0.28}
& 29.30 $\pm$ 0.45
& 36.39 $\pm$ 0.36
& 76.58 $\pm$ 0.82
& 43.48 $\pm$ 0.42 \\

FedEx-LoRA
& 30.29 $\pm$ 0.51
& 28.83 $\pm$ 0.10
& 36.57 $\pm$ 0.76
& 79.51 $\pm$ 0.25
& 45.82 $\pm$ 1.26 \\

FLoRG
& 29.83 $\pm$ 0.42
& 28.53 $\pm$ 0.54
& 36.79 $\pm$ 0.70
& 80.12 $\pm$ 0.51
& 43.48 $\pm$ 0.42 \\

FedKRSO
& 30.56 $\pm$ 0.22
& 29.40 $\pm$ 0.71
& \best{37.28 $\pm$ 0.33}
& 81.29 $\pm$ 0.49
& 45.95 $\pm$ 1.40 \\
\midrule

FedSubMuon
& 30.87 $\pm$ 0.61
& \second{29.64 $\pm$ 0.45}
& 36.49 $\pm$ 0.39
& \second{81.32 $\pm$ 0.50}
& \second{47.92 $\pm$ 1.06} \\

FedSubMuon-GT
& \best{31.37 $\pm$ 0.65}
& \best{29.70 $\pm$ 0.54}
& \second{36.86 $\pm$ 0.17}
& \best{81.79 $\pm$ 0.71}
& \best{48.44 $\pm$ 0.12} \\

\bottomrule
\end{tabular*}
\vspace{-1em}
\end{table*}

\section{Experiments}

We evaluate whether the proposed structured subspace coefficient
parameterization improves the accuracy--communication trade-off in federated
LLM fine-tuning. The experiments are designed to answer two questions: whether
\textsc{FedSubMuon} is more communication-efficient than federated LoRA and
random-subspace baselines, and whether the projected-gradient
basis tracking in \textsc{FedSubMuon-GT} translates into stronger downstream
performance. We evaluate instruction tuning and mathematical reasoning across
1B--8B LLMs.

\subsection{Setup}
\label{sec:experiment_setup}

\textbf{Models and datasets.}
We evaluate Llama 3.2 1B, Qwen3 4B, and Qwen3 8B. The instruction-tuning
benchmarks are Natural Instructions~\citep{naturalinstructions,supernaturalinstructions}
and Dolly-15K~\citep{DatabricksBlog2023DollyV2}; the reasoning benchmarks are
GSM8K~\citep{cobbe2021gsm8k} and MATH~\citep{hendrycks2021measuring}. Natural
Instructions is evaluated on Llama 3.2 1B, Dolly-15K on both Llama 3.2 1B and
Qwen3 4B, and GSM8K/MATH on Qwen3 8B.

\textbf{Evaluation metrics.}
We report ROUGE-L~\citep{lin2004rouge} for Natural Instructions and Dolly-15K,
and exact match for GSM8K and MATH. Higher values are better for both metrics.

\textbf{Baselines.}
We compare against six representative federated LoRA and random subspace baselines:
FedIT~\citep{zhang2024fedit}, FeDeRA~\citep{yan2026federa},
FLoRA~\citep{wang2024flora}, FedEx-LoRA~\citep{singhal2025fedex},
FLoRG~\citep{meng2026florg}, and FedKRSO~\citep{yang2026fedkrso},
covering factor aggregation, decomposition-based initialization,
heterogeneous-rank LoRA, exact-update correction, Gram/Procrustes aggregation,
and random subspace optimization.

\textbf{Implementation details.}
Following the LoRA experimental setting~\citep{hu2022lora}, all methods tune
only the query and value projection matrices in the attention blocks.
Training uses early stopping; therefore, the round numbers below denote the
maximum communication rounds rather than mandatory fixed training horizons.
The baselines are trained for $40$ rounds on Natural Instructions, $60$ rounds
on Dolly-15K, $20$ rounds on GSM8K, and $30$ rounds on MATH. \textsc{FedSubMuon} and
\textsc{FedSubMuon-GT} use $10$ rounds on Natural Instructions and
$30$ rounds on Dolly-15K with both backbones. On GSM8K and MATH, all
methods use the same $20$ and $30$ rounds, respectively. The shorter
\textsc{FedSubMuon} schedules on Natural Instructions and Dolly-15K highlight the
convergence advantage under fewer communication rounds, while the matched
GSM8K and MATH schedules support fair comparison on reasoning tasks. We use AdamW for the
baselines because it is a standard optimizer and directly replacing it
with Muon does not improve representative LoRA-factor baselines in our
federated setting. Natural Instructions uses task-level heterogeneity, whereas
the other datasets are partitioned with a Dirichlet distribution.
Additional hyperparameter settings are provided in Appendix~\ref{app:hyperparams},
and dataset processing details are provided in
Appendix~\ref{app:dataset_processing}.

\subsection{Performance}
\label{sec:performance}

This subsection compares the downstream performance of all methods
under the federated fine-tuning settings above.
The main goal is to test whether compact subspace coefficient
optimization can improve the accuracy of methods that communicate
larger adapter or subspace states. Table~\ref{tab:main_results_updated} reports
the task performance across datasets and model scales, and the corresponding
per-round per-client trainable parameter counts are summarized in
Appendix Table~\ref{tab:trainable_param_counts}.

Table~\ref{tab:main_results_updated} shows that across the
evaluated datasets and model scales, \textsc{FedSubMuon-GT} delivers the most
consistently strong accuracy, while \textsc{FedSubMuon} remains competitive.
\textsc{FedSubMuon-GT} is best on four out of five dataset--model pairs and second-best
on the remaining one, showing that tracked refresh improves accuracy across
both instruction tuning and reasoning settings. 

\textsc{FedSubMuon-GT} shows its
largest gains on the reasoning benchmarks. These suggest that
improving the current shared basis, rather than re-sampling it at random, is
particularly useful when stronger task adaptation is required. They are also
consistent with the difference in parameterization capacity: in conventional
LoRA methods, the trainable budget remains tied to the backbone matrix
dimensions, so the practical rank typically has to stay small as model size
grows. This pressure is especially acute in FL, where clients have limited
local compute, memory, and communication resources. This restriction can make
those baselines less expressive on more complex tasks such as mathematical
reasoning. By contrast, the \textsc{FedSubMuon} subspace coefficient
parameterization is decoupled from the full backbone matrix size, which can
allocate substantially larger ranks for harder tasks.

\subsection{Performance Analysis under Limited Communication}
\label{sec:limited_communication}

This subsection evaluates whether the accuracy advantage of \textsc{FedSubMuon} persists
after controlling for \emph{normalized communication}: the accumulated
method-dependent training traffic, counting round-wise uploads, downloads, and
refresh traffic while excluding the one-time backbone initialization payload
common to all methods. We compare all methods under the same normalized communication budget.
Table~\ref{tab:rank_budget_results} reports the results.

To instantiate the matched-budget comparison, we fix the normalized communication budget
to the corresponding \textsc{FedSubMuon} run. For the baselines, we sweep ranks
in $\{1,2,4,8\}$ and report the best result under this budget. For \textsc{FedSubMuon-GT},
the rank is set to the largest feasible even value implied by the budget.
The resulting total normalized communication budgets are $220$ MB for
NI with Llama 1B, $264$ MB for Dolly-15K with Llama 1B, and $3457$ MB for
Dolly-15K with Qwen 4B.

Even with this favorable rank-tuning opportunity for the baselines, \textsc{FedSubMuon}
is the best method in all three matched-budget settings. \textsc{FedSubMuon-GT} is also
competitive, obtaining the second-best result in two of the three settings.
FedKRSO cannot run under these budgets because its full-matrix downlink is too
large.

The results further show that simply reducing the adapter rank
increases the number of feasible updates but limits adaptation capacity, making
it an unreliable substitute for the \textsc{FedSubMuon} parameterization. \textsc{FedSubMuon}
instead preserves a larger subspace rank within the
matched budget, indicating that its subspace coefficient parameterization uses the
available communication more effectively than rank reduction alone.

\begin{table}[!t]
\centering
\small
\setlength{\tabcolsep}{4pt}
\renewcommand{\arraystretch}{1.06}
\newcommand{\rankbest}[1]{\textbf{#1}}
\newcommand{\ranksecond}[1]{\underline{#1}}
\begin{threeparttable}
\caption{Performance under matched communication budgets. Each cell reports the mean
ROUGE-L\% over two runs under the matched communication budget.}
\label{tab:rank_budget_results}
\begin{tabular}{@{}lccc@{}}
\toprule
Method & \multicolumn{3}{c}{Model / Dataset} \\
\cmidrule(l){2-4}
& Llama/NI & Llama/Dolly & Qwen/Dolly \\
\midrule
FedSubMuon & \rankbest{30.59} & \rankbest{30.09} & \rankbest{36.87} \\
FedSubMuon-GT & 29.19 & \ranksecond{29.32} & \ranksecond{36.83} \\
FedIT & \ranksecond{30.12} & 24.70 & 34.62 \\
FeDeRA & 30.10 & 26.68 & 36.49 \\
FLoRA & 10.28 & 16.95 & 28.77 \\
FedEx-LoRA & 10.24 & 16.31 & 28.02 \\
FLoRG & 29.94 & 28.98 & 35.79 \\
FedKRSO & - & - & - \\
\bottomrule
\end{tabular}
\end{threeparttable}
\end{table}

\begin{figure}[!t]
    \centering
    \begin{subfigure}[t]{0.49\linewidth}
        \centering
        \includegraphics[width=\linewidth,trim=0 0 170pt 0,clip]{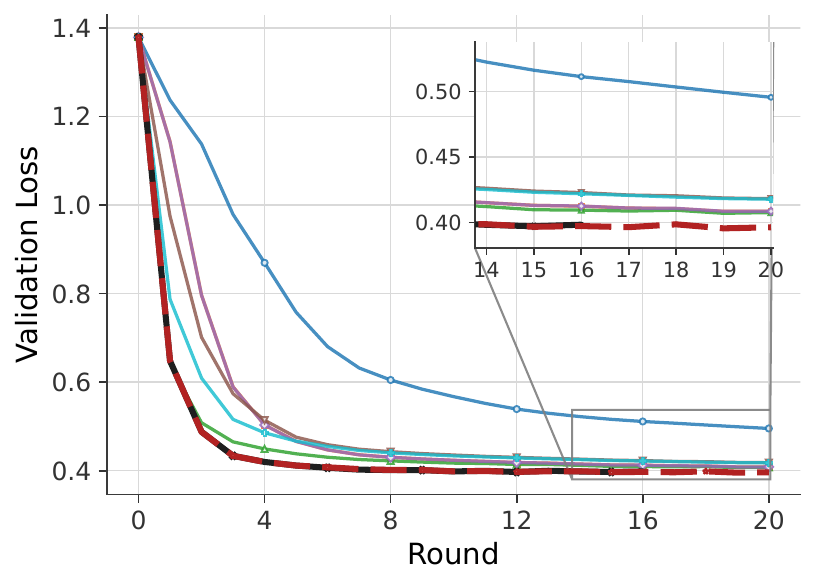}
        \caption{GSM8K, loss vs. round.}
        \label{fig:gsm8k_loss_round}
    \end{subfigure}
    \hfill
    \begin{subfigure}[t]{0.49\linewidth}
        \centering
        \includegraphics[width=\linewidth,trim=0 0 170pt 0,clip]{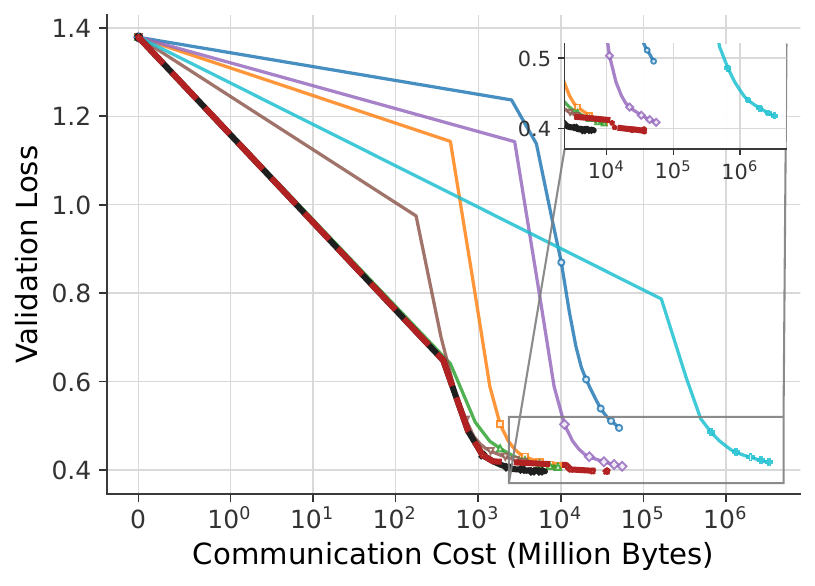}
        \caption{GSM8K, loss vs. comm.}
        \label{fig:gsm8k_loss_comm}
    \end{subfigure}
    \par\medskip
    \begin{subfigure}[t]{0.49\linewidth}
        \centering
        \includegraphics[width=\linewidth,trim=0 0 170pt 0,clip]{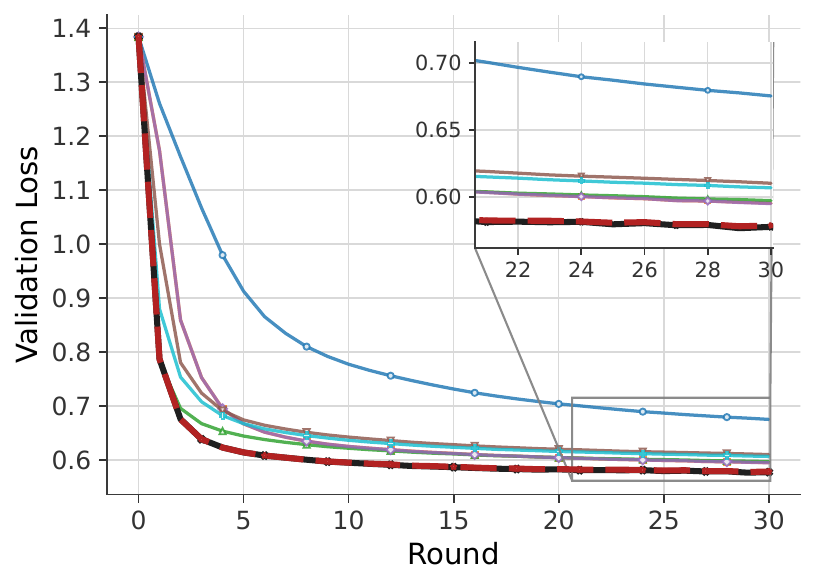}
        \caption{MATH, loss vs. round.}
        \label{fig:math_loss_round}
    \end{subfigure}
    \hfill
    \begin{subfigure}[t]{0.49\linewidth}
        \centering
        \includegraphics[width=\linewidth,trim=0 0 170pt 0,clip]{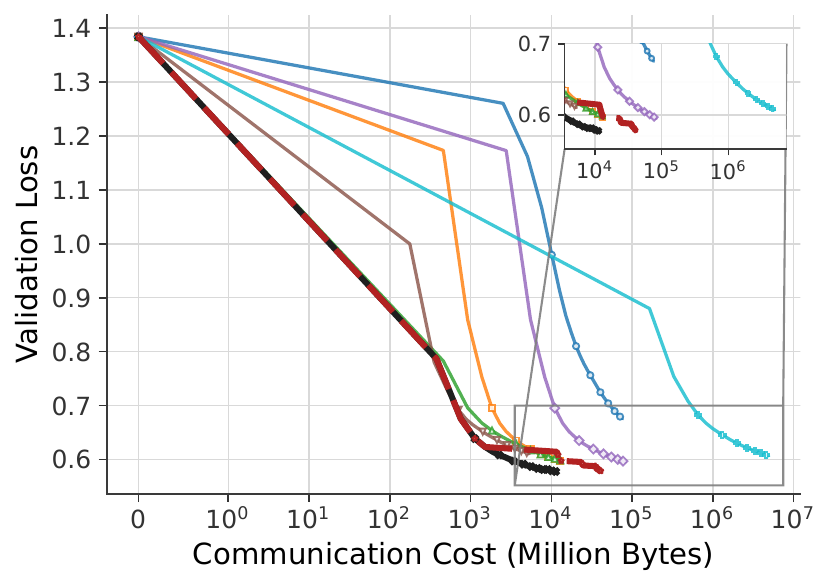}
        \caption{MATH, loss vs. comm.}
        \label{fig:math_loss_comm}
    \end{subfigure}
    \par\smallskip
    {\scriptsize
    \setlength{\tabcolsep}{2.5pt}
    \begin{tabular}{@{}llll@{}}
    \textcolor[HTML]{1F77B4}{\rule[0.5ex]{1.1em}{0.6pt}} FLoRA &
    \textcolor[HTML]{FF7F0E}{\rule[0.5ex]{1.1em}{0.6pt}} FedIT &
    \textcolor[HTML]{2CA02C}{\rule[0.5ex]{1.1em}{0.6pt}} FeDeRA &
    \textcolor[HTML]{9467BD}{\rule[0.5ex]{1.1em}{0.6pt}} FedEx-LoRA \\
    \textcolor[HTML]{8C564B}{\rule[0.5ex]{1.1em}{0.6pt}} FLoRG &
    \textcolor[HTML]{17BECF}{\rule[0.5ex]{1.1em}{0.6pt}} FedKRSO &
    \textcolor[HTML]{1F1F1F}{\rule[0.5ex]{1.1em}{0.8pt}} \textsc{FedSubMuon} &
    \textcolor[HTML]{B22222}{\rule[0.5ex]{1.1em}{0.8pt}} \textsc{FedSubMuon-GT}
    \end{tabular}}
    \caption{Validation-loss dynamics on Qwen 8B.}
    \label{fig:qwen8b_loss_curves}
    \vspace{-1em}
\end{figure}

\begin{table*}[t]
\centering
\caption{Communication and computational efficiency on Dolly-15K.}
\label{tab:total_comm_memory}
\footnotesize
\setlength{\tabcolsep}{2.4pt}
\renewcommand{\arraystretch}{1.08}

\newcommand{\best}[1]{\textbf{#1}}
\newcommand{\second}[1]{\underline{#1}}

\resizebox{\textwidth}{!}{
\begin{tabular}{llcc@{\hspace{6pt}}cccccc}
\toprule
Model & Metric
& FedSubMuon
& FedSubMuon-GT
& FedIT
& FeDeRA
& FLoRA
& FedEx-LoRA
& FLoRG
& FedKRSO \\
\midrule

\multirow{4}{*}{Llama 1B}
& Uplink (MB)
& \best{105 (1$\times$)}
& 923 (8.8$\times$)
& 920 (8.8$\times$)
& 1159 (11$\times$)
& 2045 (19$\times$)
& 920 (8.8$\times$)
& \second{721 (6.9$\times$)}
& 1311 (12$\times$) \\
& Total Comm. (MB)
& \best{264 (1$\times$)}
& 11848 (45$\times$)
& 1840 (7.0$\times$)
& 2317 (8.8$\times$)
& 22492 (85$\times$)
& 11042 (42$\times$)
& \second{1442 (5.5$\times$)}
& 85679 (325$\times$) \\
& GPU Mem. (MB)
& \best{5530}
& \second{5555}
& 5761
& 5762
& 5764
& 5762
& 6567
& 24136 \\
& Wall time (s)
& \second{77.19}
& \best{75.33}
& 83.47
& 81.25
& 82.65
& 84.61
& 114.18
& 77.61 \\
\midrule

\multirow{4}{*}{Qwen 4B}
& Uplink (MB)
& \best{1410 (1$\times$)}
& 7078 (5.0$\times$)
& 4720 (3.3$\times$)
& 5782 (4.1$\times$)
& 7078 (5.0$\times$)
& 6960 (4.9$\times$)
& \second{2490 (1.8$\times$)}
& 13920 (9.9$\times$) \\
& Total Comm. (MB)
& \best{3457 (1$\times$)}
& 83802 (24$\times$)
& 9440 (2.7$\times$)
& 11564 (3.3$\times$)
& 77880 (23$\times$)
& 83519 (24$\times$)
& \second{4980 (1.4$\times$)}
& 1127508 (326$\times$) \\
& GPU Mem. (MB)
& \best{14615}
& \second{14825}
& 15230
& 15226
& 15178
& 15230
& 23427
& 43259 \\
& Wall time (s)
& \second{100.37}
& \best{93.14}
& 108.00
& 109.89
& 111.59
& 109.11
& 234.04
& 104.50 \\
\bottomrule
\end{tabular}
}
\vspace{-1em}
\end{table*}

\subsection{Training Loss under Rounds and Communication}
\label{sec:loss_curves}

This subsection compares the training convergence behavior of all methods. We
analyze the Qwen 8B validation-loss trajectories from the preceding experiments
in two complementary views: progress per communication round and progress per
unit of normalized communication.
Figure~\ref{fig:qwen8b_loss_curves} reports the loss comparisons on GSM8K and
MATH.

\textbf{Loss versus rounds.}
The round-based curves show that \textsc{FedSubMuon} and \textsc{FedSubMuon-GT} converge faster
than the baselines and maintain a lower-loss frontier throughout training. On
GSM8K, \textsc{FedSubMuon-GT} reaches the lowest final validation loss at $0.396$,
followed closely by \textsc{FedSubMuon} at $0.397$, while both enter the low-loss region
earlier than the strongest non-\textsc{FedSubMuon} baselines. On MATH, \textsc{FedSubMuon} and
\textsc{FedSubMuon-GT} are essentially tied, ending at $0.577$ and $0.578$,
respectively, and both remain below the baseline curves. These trajectories
therefore show that the proposed subspace parameterization improves not only
the final scores in Table~\ref{tab:main_results_updated}, but also the speed and
quality of optimization during training.

\textbf{Loss versus communication.}
The plots examine how validation loss decreases as
cumulative normalized communication grows. \textsc{FedSubMuon} traces the
strongest lower-left frontier on both datasets, reaching the low-loss region with
substantially less traffic than the LoRA baselines. FedKRSO also
shows a pronounced initial loss drop, but each round communicates more
parameters than the LoRA baselines, shifting its curve to the right on the
communication axis.
\textsc{FedSubMuon-GT} shifts this frontier by spending extra basis-refresh communication
on tracked subspace bases. This places \textsc{FedSubMuon-GT} to the right of \textsc{FedSubMuon} in
the communication-axis panels, reflecting the cost of basis tracking.

\subsection{Communication and Computational Efficiency}
\label{sec:efficiency}

This subsection analyzes Table~\ref{tab:total_comm_memory} to compare the
communication and computational efficiency of the proposed methods against the
baselines. All measurements use the same Dolly-15K settings as
Table~\ref{tab:main_results_updated}. The total-communication entries report
normalized communication, with parenthetical ratios relative to
\textsc{FedSubMuon} for each backbone; uplink traffic is reported separately.
GPU memory denotes the peak footprint, and wall time denotes the average
per-round runtime.

\textbf{Communication efficiency.}
\textsc{FedSubMuon} is the clear communication minimum on both backbones. The closest alternative is
FLoRG, which still needs $5.5\times$ more communication on Llama 1B and
$1.4\times$ more on Qwen 4B. The LoRA-style baselines and FedKRSO require
substantially larger budgets; for example, FedIT uses $7.0\times$ and
$2.7\times$ more communication than \textsc{FedSubMuon} on the two backbones, while
FedKRSO uses $325\times$ and $326\times$ more because it requires broadcasting
a full selected-layer matrix update in each round.
\textsc{FedSubMuon-GT} spends additional communication on gradient-tracked subspace refreshes
and lazy catch-up packages ($45\times$ on Llama 1B and $24\times$ on Qwen 4B),
but this extra budget is
paired with the strongest overall accuracy. Importantly, most of this overhead
is not uplink traffic: \textsc{FedSubMuon-GT} uploads $923$ MB on Llama 1B and
$7078$ MB on Qwen 4B, only $7.8\%$ and $8.4\%$ of its total communication.
This matters in cross-device FL, where uplink and downlink resources are
asymmetric and uplink is often the tighter client-side bottleneck.

\textbf{Computational efficiency.}
The computation-side metrics show that the stronger accuracy of \textsc{FedSubMuon-GT}
does not come with heavy runtime or memory overhead. \textsc{FedSubMuon} and
\textsc{FedSubMuon-GT} achieve the best and second-best GPU memory on both backbones,
and \textsc{FedSubMuon-GT} is the fastest method in wall-clock time, with \textsc{FedSubMuon}
ranking second.

\subsection{Ablation Studies}
\label{sec:ablation_studies}

\begin{figure}[t]
    \centering
    \begin{subfigure}[t]{0.49\linewidth}
        \centering
        \includegraphics[width=\linewidth]{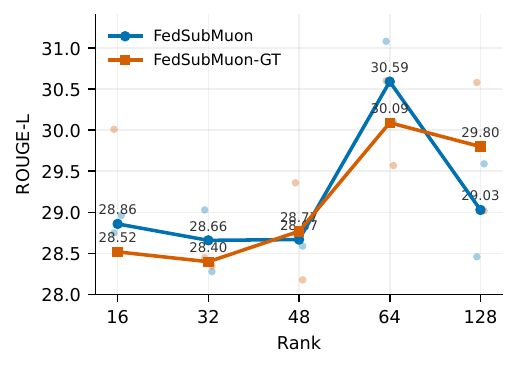}
        \caption{Rank sweep.}
        \label{fig:ablation_rank}
    \end{subfigure}
    \hfill
    \begin{subfigure}[t]{0.49\linewidth}
        \centering
        \includegraphics[width=\linewidth]{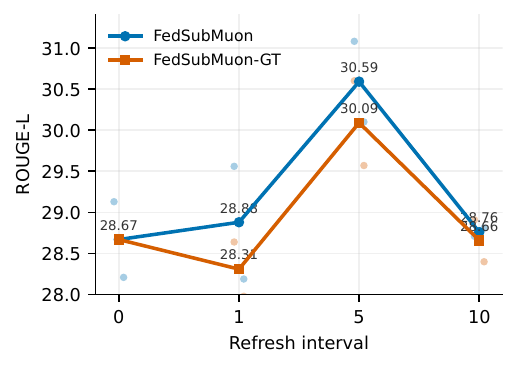}
        \caption{Refresh-interval sweep.}
        \label{fig:ablation_refresh}
    \end{subfigure}
    \par\smallskip
    \begin{subfigure}[t]{0.49\linewidth}
        \centering
        \includegraphics[width=\linewidth]{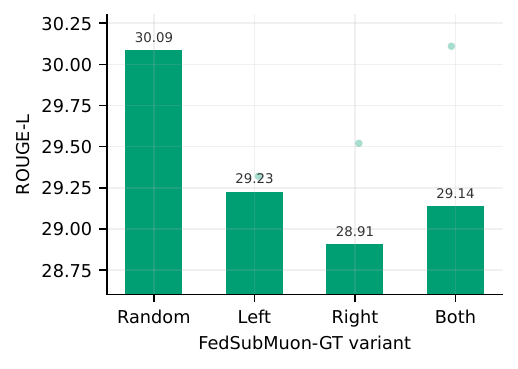}
        \caption{Basis initialization variants for \textsc{FedSubMuon-GT}.}
        \label{fig:ablation_svd}
    \end{subfigure}
    \hfill
    \begin{subfigure}[t]{0.49\linewidth}
        \centering
        \includegraphics[width=\linewidth]{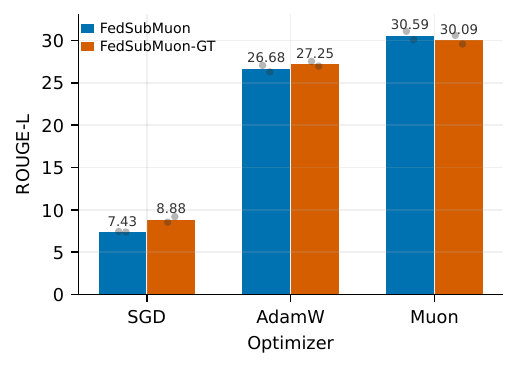}
        \caption{Optimizer sweep.}
        \label{fig:ablation_optimizer}
    \end{subfigure}
    \caption{\textsc{FedSubMuon} component ablations on Dolly-15K with Llama 1B. Each plotted
    value is the mean over two runs.}
    \label{fig:dolly_llama1b_ablation}
\end{figure}

This section isolates the internal design behind the \textsc{FedSubMuon} and
\textsc{FedSubMuon-GT} configurations, including subspace rank, basis refresh, basis
initialization, and optimizer choice.
Appendix~\ref{app:ablation_studies} additionally varies the federation setting.
The results in Figure~\ref{fig:dolly_llama1b_ablation} support four conclusions:
(1) moderate subspace ranks are sufficient; (2) basis refresh should be used
with a reasonable frequency; (3) random basis initialization outperforms SVD-based alternatives;
and (4) Muon is important for optimizing the subspace coefficient matrix.

\textbf{Rank.}
Figure~\ref{fig:ablation_rank} shows that increasing the subspace rank improves
performance up to a point but does not monotonically improve.
\textsc{FedSubMuon} reaches its best score at rank $64$ ($30.59$), while
\textsc{FedSubMuon-GT} also performs best at rank $64$ ($30.09$) and remains competitive
at rank $128$ ($29.80$). Lower ranks underfit the update space, whereas the
rank-$128$ setting does not provide a clear gain on this backbone.

\textbf{Basis refresh.}
Figure~\ref{fig:ablation_refresh} evaluates the refresh interval used by
\textsc{FedSubMuon-GT}. When the refresh interval is zero, \textsc{FedSubMuon-GT} performs the
same computation as \textsc{FedSubMuon} because the basis is never refreshed. Refreshing every five rounds gives the
best \textsc{FedSubMuon-GT} result ($30.09$), while refreshing too frequently
($\tau=1$) or too rarely ($\tau=10$) reduces performance to $28.31$ and $28.66$,
respectively. This pattern indicates that basis refresh is useful when it
periodically expands the explored subspace, but overly frequent refreshes can
destabilize local optimization.

\textbf{SVD initialization.}
Figure~\ref{fig:ablation_svd} compares the default random initialization used
by \textsc{FedSubMuon-GT} against SVD-based initialization variants. Instead of drawing
all subspace bases from random seeds, the SVD variants initialize one or both
bases from the singular vectors of the frozen pretrained matrix $W_\ell^0$.
Concretely, if
$W_\ell^0 \approx \widetilde U_\ell \Sigma_\ell \widetilde V_\ell^\top$, the
left variant initializes $U_\ell^0$ with the leading columns of
$\widetilde U_\ell$ while keeping $V_\ell^0$ random, the right variant
initializes $V_\ell^0$ with the leading columns of $\widetilde V_\ell$ while
keeping $U_\ell^0$ random, and the both variant initializes both sides from the
corresponding singular-vector bases. This test follows the same motivation as
SVD-based initializations, which replace random adapter directions with
principal weight or activation subspaces~\citep{yan2026federa,meng2024pissa,paischer2025parameter}.
The random initialization gives the highest score overall ($30.09$), exceeding
the left ($29.23$), both ($29.14$), and right ($28.91$) variants. Among the
three SVD-based variants, initializing the left basis performs best. This
result indicates that, in this setting, the default random subspace is more
effective than replacing it with singular-vector initialization.

\textbf{Optimizer.}
Figure~\ref{fig:ablation_optimizer} shows that the optimizer choice has the
largest effect. SGD fails to make meaningful progress, reaching only $7.43$ for
\textsc{FedSubMuon} and $8.88$ for \textsc{FedSubMuon-GT}. AdamW is substantially better
($26.68$ and $27.25$), but still trails Muon by $3.91$ and $2.84$ ROUGE-L
points. Muon therefore provides a critical optimization advantage for the
subspace coefficient matrix, which explains why the main experiments use Muon
for both \textsc{FedSubMuon} variants.

\section{Conclusion}

We introduced \textsc{FedSubMuon}, a communication-efficient realization of
federated Muon that moves matrix-aware optimization from full weight matrices
to compact structured subspace coefficients. This design keeps Muon on a
single matrix-valued trainable object and reduces client upload to compact coefficient matrices.
Empirically, while \textsc{FedSubMuon} is the strongest communication-efficient
variant, \textsc{FedSubMuon-GT} further adds
projected-gradient basis tracking and achieves the strongest overall accuracy. These results show that
structured subspace coefficients provide an effective path for bringing
Muon matrix optimization to federated LLM fine-tuning.

\section*{Limitations}

The primary limitation of this work is that \textsc{FedSubMuon} is evaluated
in controlled federated fine-tuning settings rather than in a real-world
cross-device deployment. Although our experiments cover instruction tuning and
reasoning tasks with 1B--8B LLMs, practical deployments may involve larger
models, more heterogeneous client hardware, unstable network conditions,
stragglers, and stricter memory constraints. These factors can affect the
efficiency of client synchronization, basis refresh, and update aggregation.
We leave large-scale deployment studies with more diverse client environments
and larger foundation models to future work.

Another limitation is that our experiments focus on adapting selected attention
projection matrices. This setting follows common PEFT practice for LLM
fine-tuning, but it does not exhaust all possible adaptation targets or model
architectures. Other modules, larger adaptation scopes, or different backbone
families may exhibit different accuracy--communication trade-offs. We believe
the design principle of optimizing compact matrix-valued subspace
coefficients is broadly applicable, but a systematic exploration of broader
adaptation targets and model scales remains future work.

Finally, federated learning keeps raw examples decentralized, but it does not
by itself provide a formal privacy guarantee. Model updates, coefficient
matrices, or gradient-related information may still leak information under
strong attacks. Practical deployments on sensitive data should combine methods
such as \textsc{FedSubMuon} with appropriate privacy and security mechanisms,
such as secure aggregation and differential privacy.

\section*{Ethical Considerations}

We propose a communication-efficient federated fine-tuning framework that
reduces communication cost while keeping raw client data decentralized. All
experiments in this paper use public benchmark datasets, and we do not collect
new personal data or conduct human-subject studies. We have not identified
specific additional risks arising from the proposed method beyond those
commonly associated with federated learning and LLM fine-tuning.

\section*{Acknowledgments}

Shuzhen Chen is supported in part by the Fundamental Research Funds for the Central Universities under Grant 202513024, in part by the Qingdao Postdoctoral Research Project under Grant No. QDBSH20250102012.
Di Wang is supported in part by the funding BAS/1/1689-01-01, RGC/3/7125-01-01, FCC/1/5940-20-05, FCC/1/5940-06-02, and King Abdullah University of Science and Technology (KAUST) -- Center of Excellence for Generative AI, under award number 5940 and a gift from Google.

\bibliography{refs}

\appendix

\section{Related Work}

We situate \textsc{FedSubMuon} relative to three lines of work:
communication-efficient federated fine-tuning, random subspace
optimization, and matrix-aware optimizers.

\subsection{Federated Fine-Tuning}

Adapting LLMs under federated constraints requires minimizing
per-round communication while preserving adaptation quality.
Most approaches combine federated averaging with
parameter-efficient tuning to reduce the communicated
state~\citep{zhang2023fedpetuning,zhang2024fedit,tao2026byzantine,chen2026wireless,tao2024communication,xiang2023practical}.

LoRA-based methods communicate factorized adapter states.
FeDeRA initializes adapters from pretrained weight
decomposition~\citep{yan2026federa};
FLoRA supports heterogeneous per-client
ranks~\citep{wang2024flora};
FedEx-LoRA corrects the aggregation mismatch that arises from
separately averaging low-rank factors~\citep{singhal2025fedex}.
A complementary line replaces factorized adapters with compact
subspace coordinates generated from shared random seeds:
FLoRG communicates Gram-matrix subspace
coordinates~\citep{meng2026florg}, and FedKRSO uses
random subspace projections with memory-efficient
aggregation~\citep{yang2026fedkrso}.

\textsc{FedSubMuon} is most closely related to this seeded-subspace
group, but differs in two respects: the server maintains an
ambient-space matrix accumulator rather than a low-rank adapter,
and clients apply Muon to a matrix-valued coefficient
matrix rather than optimizing scalar subspace coordinates.

\subsection{Random Subspace Optimization}

The intrinsic-dimension framework establishes that neural networks
can reach competitive task performance when optimization is
restricted to a low-dimensional random
subspace~\citep{li2018intrinsic}.
For LLMs, this premise extends to fine-tuning: effective adaptation
can be achieved in surprisingly small subspaces of the pretrained
parameter space~\citep{aghajanyan2021intrinsic}, and empirical
analyses show that fine-tuning trajectories concentrate on compact
task-relevant directions~\citep{zhang2023tiny}.
LoRA can be understood as a structured instantiation of this
principle, tying the trainable budget to a low-rank
factorization~\citep{hu2022lora,ding2023delta}.

\textsc{FedSubMuon} uses the low-dimensionality premise differently
from standard LoRA\@.
The subspace coefficient matrix is a per-round optimization variable
rather than a persistent adapter: committed updates accumulate in
the ambient matrix, so refreshing or tracking the bases
redirects future search without discarding previously accumulated
gradient information.

\subsection{Muon and Matrix-Aware Optimizers}

Matrix-aware optimizers exploit structure invisible to
coordinate-wise methods.
Shampoo maintains per-dimension Kronecker-factor
preconditioners~\citep{gupta2018shampoo};
Sophia incorporates lightweight diagonal Hessian
estimates~\citep{liu2024sophia};
SOAP stabilizes Shampoo-style preconditioning by maintaining Adam
moments in the preconditioner eigenbasis~\citep{vyas2025soap}.
These methods improve optimization geometry but require
maintaining and computing preconditioning matrices that introduce
memory and computational overhead beyond the weights themselves.

Muon offers a lighter alternative by orthogonalizing momentum via
Newton--Schulz iterations and applying the resulting matrix
direction to hidden-layer weights~\citep{jordan2024muon}.
Recent work on federated Muon analyzes convergence and
acceleration under matrix-orthogonalized
updates~\citep{liu2025fedmuon,takezawa2025fedmuon,%
zhang2025provablefedmuon}, but these methods operate on full weight
matrices, incurring full layer-size communication per round.
Applying Muon directly to LoRA factors is not equivalent to
optimizing the underlying update matrix, because the factors are
non-unique coordinates of a low-rank
product~\citep{bogachev2026riemannion}.
\textsc{FedSubMuon} resolves this by applying Muon to a
$r \times r$ subspace coefficient matrix, preserving Muon's
matrix-update geometry while reducing the federated payload to
$O(r^2)$ per layer.

\section{Algorithms}
\label{app:unified_algorithm}

Algorithms~\ref{alg:fedsubmuon} and~\ref{alg:fedsubmuon_gt} summarize
\textsc{FedSubMuon} and \textsc{FedSubMuon-GT}, respectively.

\begin{algorithm}[!t]
\caption{\algcaptionyellow{FedSubMuon}}
\label{alg:fedsubmuon}
\scriptsize
\setlength{\abovedisplayskip}{1pt}
\setlength{\belowdisplayskip}{1pt}
\setlength{\abovedisplayshortskip}{1pt}
\setlength{\belowdisplayshortskip}{1pt}
\begin{algorithmic}[1]
\Require \algwrap{Backbone $W^0$, rounds $T$, local steps $E$, client weights $\{p_i\}_{i=1}^N$, selected-matrix index set $\mathcal L$, subspace rank $r$, refresh interval $\tau$, learning rate $\eta$}
\StateW{Initialize global update accumulators $\Delta W_\ell^0\gets0$, local versions $v_i\gets0$, local update accumulators $\Delta W_{i,\ell}\gets0$, trainable subspace coefficient matrices $X_\ell^0\gets0$, random seeds $s^0$, commit log $\mathcal H\gets[\,]$, counter $K\gets0$}
\For{$t=0,\ldots,T-1$}
    \StateW{\textbf{Server: }Sample clients $\mathcal S_t$ and set $a_i^t=p_i/\sum_{j\in\mathcal S_t}p_j$; send $\{s^t,X^t\}$ and missing commit records $\{\mathcal C_k\}_{k=v_i}^{K-1}$ to each $i\in\mathcal S_t$}
    \Statex \textit{// Client synchronization}
    \ForAll{$i\in\mathcal S_t$ \textbf{in parallel}}
        \StateW{For each $\mathcal C_k$, generate $(U_\ell^{[k]},V_\ell^{[k]})$ from $s^{[k]}$ and apply $\Delta W_{i,\ell}\gets\Delta W_{i,\ell}+U_\ell^{[k]}X_\ell^{[k]}(V_\ell^{[k]})^\top$ for all $\ell\in\mathcal L$; set $v_i\gets K$}
        \StateW{Generate $(U_\ell^t,V_\ell^t)$ from seed $s^t$;\\ Initialize $X_{i,\ell}^{t,0}$, $M_{i,\ell}^{t,0}$, $W_{i,\ell}(X_{i,\ell}^{t,0})$ by Eqs.~\eqref{eq:core_lift}--\eqref{eq:core_init}}
        \Statex \textit{// Local Muon update of the subspace coefficient matrix}
        \StateW{Run $E$ local Muon steps on $X_{i,\ell}$ using Eqs.~\eqref{eq:core_gradient}--\eqref{eq:core_muon_update};\\set $X_{i,\ell}^{t+1}\gets X_{i,\ell}^{t,E}$}
        \StateYellow{Upload $X_{i,\ell}^{t+1}$ for all $\ell\in\mathcal L$}
    \EndFor
    \Statex \textit{// Server aggregation}
    \StateW{Set $\bar X_\ell^{t+1}\gets\sum_{i\in\mathcal S_t}a_i^tX_{i,\ell}^{t+1}$ for all $\ell\in\mathcal L$}
    \Statex \textit{// Refresh}
    \IfYellow{$(t+1)\bmod\tau=0$}
        \StateYellow{$\mathcal C_K\gets(s^{[K]}=s^t,\{X_\ell^{[K]}=\bar X_\ell^{t+1}\}_{\ell\in\mathcal L})$\\ Append $\mathcal C_K$ to $\mathcal H$; set $K\gets K+1$}
        \StateYellow{$\Delta W_\ell^{t+1}\gets\Delta W_\ell^t+U_\ell^t\bar X_\ell^{t+1}(V_\ell^t)^\top$ for all $\ell\in\mathcal L$; set $X_\ell^{t+1}\gets0$}
        \StateYellow{Sample fresh random seeds $s^{t+1}$}
    \ElseYellow
        \StateYellow{Keep $\Delta W_\ell^{t+1}\gets\Delta W_\ell^t$, $s^{t+1}\gets s^t$\\ Set $X_\ell^{t+1}\gets\bar X_\ell^{t+1}$}
    \EndIf
\EndFor
\StateW{\Return $W^0+\Delta W_\ell^T+U_\ell^TX_\ell^T(V_\ell^T)^\top$}
\end{algorithmic}
\end{algorithm}

\begin{algorithm}[!t]
\caption{\algcaptionblue{FedSubMuon-GT}}
\label{alg:fedsubmuon_gt}
\scriptsize
\setlength{\abovedisplayskip}{1pt}
\setlength{\belowdisplayskip}{1pt}
\setlength{\abovedisplayshortskip}{1pt}
\setlength{\belowdisplayshortskip}{1pt}
\begin{algorithmic}[1]
\Require \algwrap{Backbone $W^0$, rounds $T$, local steps $E$, client weights $\{p_i\}_{i=1}^N$, selected-matrix index set $\mathcal L$, subspace rank $r$, refresh interval $\tau$, learning rate $\eta$, subspace learning rate $\rho$}
\StateW{Initialize global update accumulators $\Delta W_\ell^0\gets0$, local versions $v_i\gets0$, local update accumulators $\Delta W_{i,\ell}\gets0$, trainable subspace coefficient matrices $X_\ell^0\gets0$, random seed $s^0$ for the initial subspace, commit log $\mathcal H\gets[\,]$, counter $K\gets0$}
\For{$t=0,\ldots,T-1$}
    \StateW{\textbf{Server: }Sample clients $\mathcal S_t$ and set $a_i^t=p_i/\sum_{j\in\mathcal S_t}p_j$; send $\{s^t,X^t\}$ if $t<\tau$, otherwise send $\{U^t,V^t,X^t\}$, and send missing commit records $\{\mathcal C_k\}_{k=v_i}^{K-1}$ to each $i\in\mathcal S_t$}
    \Statex \textit{// Client synchronization}
    \ForAll{$i\in\mathcal S_t$ \textbf{in parallel}}
        \StateW{For each $\mathcal C_k$, apply $\Delta W_{i,\ell}\gets\Delta W_{i,\ell}+U_\ell^{[k]}X_\ell^{[k]}(V_\ell^{[k]})^\top$ for all $\ell\in\mathcal L$; set $v_i\gets K$}
        \StateW{Generate $(U_\ell^t,V_\ell^t)$ from seed $s^t$ if $t<\tau$; otherwise use received tracked subspace bases $(U_\ell^t,V_\ell^t)$;\\ Initialize $X_{i,\ell}^{t,0}$, $M_{i,\ell}^{t,0}$, $W_{i,\ell}(X_{i,\ell}^{t,0})$ by Eqs.~\eqref{eq:core_lift}--\eqref{eq:core_init}}
        \Statex \textit{// Local Muon update of the subspace coefficient matrix}
        \StateW{Run $E$ local Muon steps on $X_{i,\ell}$ using Eqs.~\eqref{eq:core_gradient}--\eqref{eq:core_muon_update};\\set $X_{i,\ell}^{t+1}\gets X_{i,\ell}^{t,E}$}
        \IfBlue{$(t+1)\bmod\tau=0$}
            \StateBlue{Form projected gradients by Eq.~\eqref{eq:gt_gradient_summaries}}
            \StateBlue{Upload $\{X_{i,\ell}^{t+1},H_{U,i,\ell}^t,H_{V,i,\ell}^t\}_{\ell\in\mathcal L}$}
        \ElseBlue
            \StateBlue{Upload $\{X_{i,\ell}^{t+1}\}_{\ell\in\mathcal L}$}
        \EndIf
    \EndFor
    \Statex \textit{// Server aggregation}
    \StateW{Set $\bar X_\ell^{t+1}\gets\sum_{i\in\mathcal S_t}a_i^tX_{i,\ell}^{t+1}$ for all $\ell\in\mathcal L$}
    \Statex \textit{// Refresh}
    \IfBlue{$(t+1)\bmod\tau=0$}
        \StateBlue{$\mathcal C_K\gets(\{U_\ell^{[K]}=U_\ell^t,V_\ell^{[K]}=V_\ell^t,X_\ell^{[K]}=\bar X_\ell^{t+1}\}_{\ell\in\mathcal L})$; Append $\mathcal C_K$ to $\mathcal H$; set $K\gets K+1$}
        \StateBlue{$\Delta W_\ell^{t+1}\gets\Delta W_\ell^t+U_\ell^t\bar X_\ell^{t+1}(V_\ell^t)^\top$ for all $\ell\in\mathcal L$; set $X_\ell^{t+1}\gets0$}
        \StateBlue{Aggregate projected gradients by Eq.~\eqref{eq:gt_summary_aggregate}; form least-squares coordinates and tangent directions by Eqs.~\eqref{eq:gt_ls_coordinates}--\eqref{eq:gt_residual_direction}; update $(U_\ell^{t+1},V_\ell^{t+1})$ by Eq.~\eqref{eq:gt_basis_retraction}}
    \ElseBlue
        \StateBlue{Keep $\Delta W_\ell^{t+1}\gets\Delta W_\ell^t$, keep $s^{t+1}\gets s^t$ if $t+1<\tau$ and $(U_\ell^{t+1},V_\ell^{t+1})\gets(U_\ell^t,V_\ell^t)$ otherwise, and set $X_\ell^{t+1}\gets\bar X_\ell^{t+1}$}
    \EndIf
\EndFor
\StateW{\Return $W^0+\Delta W_\ell^T+U_\ell^TX_\ell^T(V_\ell^T)^\top$}
\end{algorithmic}
\end{algorithm}

\section{Hyperparameter Settings}
\label{app:hyperparams}

For Natural Instructions, we use $738$ clients and sample $3\%$ of clients per
communication round. For Dolly-15K, we use $200$ clients with $5\%$ participation.
For GSM8K and MATH, we use $100$ clients with $10\%$ participation.

\begin{table}[t]
\centering
\scriptsize
\setlength{\tabcolsep}{2pt}
\renewcommand{\arraystretch}{1.04}
\begin{threeparttable}
\caption{Per-round per-client trainable parameter counts under the $q/v$-only
adaptation setup.}
\label{tab:trainable_param_counts}
\begin{tabularx}{\columnwidth}{@{}l
    >{\raggedright\arraybackslash}X
    c
    >{\raggedleft\arraybackslash}X@{}}
\toprule
\textbf{Model} &
\textbf{Method group} &
\textbf{Rank} &
\textbf{Trainable params (ratio)} \\
\midrule
Llama 1B
& FedSubMuon/\newline FedSubMuon-GT
& $64$
& $131{,}072$ ($1\times$) \\
\cmidrule(lr){2-4}
& FedIT/FeDeRA/FLoRA/\newline FedEx-LoRA
& $8$
& $851{,}968$ ($6.5\times$) \\
& FLoRG
& $8$
& $327{,}680$ ($2.5\times$) \\
& FedKRSO
& $8$
& $1{,}310{,}720$ ($10\times$) \\
\midrule
Qwen 4B
& FedSubMuon/\newline FedSubMuon-GT
& $128$
& $1{,}179{,}648$ ($1\times$) \\
\cmidrule(lr){2-4}
& FedIT/FeDeRA/FLoRA/\newline FedEx-LoRA
& $8$
& $2{,}949{,}120$ ($2.5\times$) \\
& FLoRG
& $8$
& $1{,}032{,}192$ ($0.88\times$) \\
& FedKRSO
& $8$
& $5{,}898{,}240$ ($5\times$) \\
\midrule
Qwen 8B
& FedSubMuon/\newline FedSubMuon-GT
& $256$
& $4{,}718{,}592$ ($1\times$) \\
\cmidrule(lr){2-4}
& FedIT/FeDeRA/FLoRA/\newline FedEx-LoRA
& $12$
& $5{,}750{,}784$ ($1.22\times$) \\
& FLoRG
& $12$
& $2{,}211{,}840$ ($0.47\times$) \\
& FedKRSO
& $12$
& $8{,}847{,}360$ ($1.88\times$) \\
\bottomrule
\end{tabularx}
\end{threeparttable}
\end{table}

For all experiments, each participating client uses a local batch size of $1$.
All inputs are tokenized with a maximum length of $1024$. Each client runs
$100$ local optimization steps on Natural Instructions and one local epoch per
communication round on the remaining datasets. We apply early stopping with
patience $5$ and min-delta $8$.For the baseline methods, we use
AdamW and search the learning rate over
$\{10^{-4}, 5\times 10^{-5}, 10^{-5}, 5\times 10^{-6}\}$. The final learning
rate is $10^{-5}$ for FedKRSO and $5\times 10^{-5}$ for all other baselines.
For all methods, we set the LoRA scaling factor to $1$.

For \textsc{FedSubMuon} and \textsc{FedSubMuon-GT}, we sweep the coefficient-matrix learning rate over
$\{10^{-3}, 2\times 10^{-3}, 5\times 10^{-3}, 10^{-2}\}$. \textsc{FedSubMuon} uses a
final coefficient-matrix learning rate of $2\times 10^{-3}$ on Natural Instructions and
$5\times 10^{-3}$ on the remaining datasets. For
\textsc{FedSubMuon-GT}, we additionally search the subspace learning rate over
$\{5\times 10^{-3}, 10^{-2}, 5\times 10^{-2}, 10^{-1}\}$. On Natural
Instructions, the final hyperparameters are a coefficient-matrix learning rate of
$2\times 10^{-3}$ and a subspace learning rate of $10^{-2}$. On the remaining
datasets, \textsc{FedSubMuon-GT} uses a coefficient-matrix learning rate of $5\times 10^{-3}$ and a
subspace learning rate of $5\times 10^{-2}$.

\section{Dataset Processing and Prompt Templates}
\label{app:dataset_processing}

This section summarizes the dataset preprocessing pipeline used in our codebase.
For the Qwen backbones, each example is formatted as a single-turn chat request and then
processed with the model's chat template. For the non-Qwen backbone, we use
plain-text instruction prompts.

\textbf{Dolly-15K.}
Each example is normalized into an instruction, an optional context field used
as input, a response target, and the original category label. The prompt for
Llama follows the standard Alpaca format with an instruction block, an optional
input block when the context is non-empty, and a response header. For Qwen,
the same content is converted into a user message containing the instruction
followed by the optional input field before applying the chat template.
Dolly-15K contains eight task categories: brainstorming, classification, closed
QA, creative writing, general QA, information extraction, open QA, and
summarization. In our split, the summarization category is used as the
evaluation set, and the remaining seven categories form the federated training
set, which is then partitioned across clients according to the specified IID or
Dirichlet setting. The default partition is Dirichlet with concentration $0.5$.

\textbf{Natural Instructions.}
Each example is converted into a triple consisting of the task definition, the
instance input, and the first reference output. For the Llama backbone, the
task definition and instance input are formatted as plain-text instructions; for
Qwen, the same content is converted into an equivalent single-turn chat
request. Beyond token truncation, we also discard examples whose raw input text
is excessively long before tokenization. After filtering, training tasks with
fewer than $20$ examples are removed; each retained training task is then
subsampled to $20\%$ of its examples; and each evaluation task is subsampled to
$\max(20, 2\%)$ examples when it contains more than $20$ instances. In the
federated split, each retained training task is treated as one client.

\textbf{GSM8K.}
Each sample is reformatted into the fixed instruction ``Solve the following
grade-school math problem step by step,'' with the question as input and the
annotated answer as target output. For the Llama backbone, the resulting sample
is formatted as a plain-text instruction example, while Qwen receives the
corresponding single-turn chat request. The official training split is
partitioned for federated training, and the official test split is used for
both eval and test. Evaluation is measured by exact-match accuracy on the
final answer.

\textbf{MATH.}
Each problem is paired with the fixed instruction ``Solve the following
competition math problem. Show your reasoning clearly and put the final answer
in \texttt{\textbackslash boxed\{\}},'' using the problem statement as input and
the full solution as target output. The preprocessing also stores the extracted
boxed final answer, together with the subject and difficulty-level metadata. As
with GSM8K, the resulting sample is formatted as plain-text instruction data
for Llama and as a single-turn chat request for Qwen. For evaluation, the
official training split is divided into a federated training partition and a
development set of size $500$; the development set is used for evaluation, and the
official test split is used for test. Performance is measured by exact-match
accuracy on the extracted boxed answer.

\section{Additional Ablation Studies}
\label{app:ablation_studies}
\label{sec:controlled_federation_sweeps}

\begin{figure}[t]
    \centering
    \includegraphics[width=\linewidth]{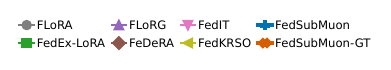}
    \par\smallskip
    \begin{subfigure}[t]{0.49\linewidth}
        \centering
        \includegraphics[width=\linewidth]{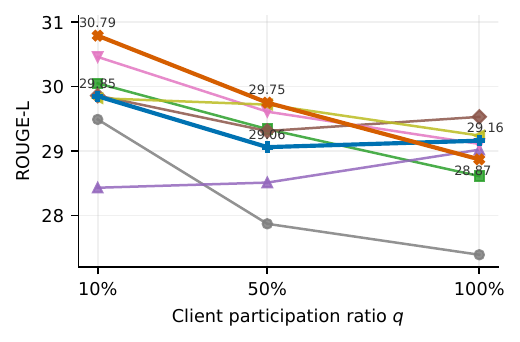}
        \caption{Client participation ratio.}
        \label{fig:ablation_participation}
    \end{subfigure}
    \hfill
    \begin{subfigure}[t]{0.49\linewidth}
        \centering
        \includegraphics[width=\linewidth]{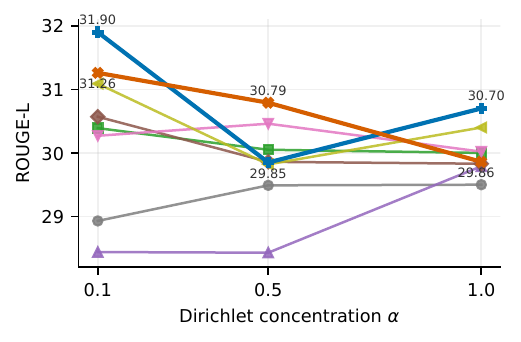}
        \caption{Client-data distribution.}
        \label{fig:ablation_distribution}
    \end{subfigure}
    \caption{Federation ablations over client participation and data
    heterogeneity on Dolly-15K with Llama 1B; each plotted value is averaged
    over two runs.}
    \label{fig:ablation_controlled_sweeps}
    \vspace{-1em}
\end{figure}

This section evaluates whether the \textsc{FedSubMuon} variants remain robust when
the federation setting changes, complementing the component ablations in
Section~\ref{sec:ablation_studies}. Figure~\ref{fig:ablation_controlled_sweeps}
varies client participation and client-data heterogeneity on Dolly-15K with
Llama 1B, showing how the methods behave under different partial-participation
and non-IID regimes. Both panels in
Figure~\ref{fig:ablation_controlled_sweeps} use $20$ total clients. In
Figure~\ref{fig:ablation_participation}, participation ratios $q=0.1$, $0.5$,
and $1.0$ correspond to low, medium, and full participation, respectively, with fixed $\alpha=0.5$. In
Figure~\ref{fig:ablation_distribution}, the Dirichlet concentration
$\alpha$ controls data heterogeneity while $q$ is fixed to $0.1$; smaller $\alpha$ indicates a stronger
non-IID split.

\textbf{Client participation ratio.}
Figure~\ref{fig:ablation_participation} sweeps the client participation ratio
across partial and full participation. \textsc{FedSubMuon-GT} achieves the best result
at low and medium participation, reaching $30.79$ at $q=0.1$ and $29.75$ at
$q=0.5$. Relative to \textsc{FedSubMuon}, the gain from basis
refresh is also large under partial participation: $+0.94$ at $q=0.1$ and
$+0.69$ at $q=0.5$. At full participation, however, most methods perform
similarly. This pattern suggests that the refreshed subspace is most useful in
the partial-participation regime targeted by cross-device federated learning,
whereas under full participation the aggregation problem becomes easier and the
methods become more comparable.

\textbf{Client-data distribution.}
Figure~\ref{fig:ablation_distribution} sweeps the Dirichlet concentration in
the fixed-participation setting. \textsc{FedSubMuon} is the strongest method at both
ends of the heterogeneity range, reaching $31.90$ at $\alpha=0.1$ and $30.70$
at $\alpha=1.0$. These scores exceed the strongest baseline by $1.33$ and
$0.30$ ROUGE-L, respectively. \textsc{FedSubMuon-GT} instead peaks at the intermediate
default setting $\alpha=0.5$, where it reaches $30.79$ and improves over the
best baseline by $0.33$ ROUGE-L. Together, these results show that the
\textsc{FedSubMuon} variants remain competitive across most client-data heterogeneity
settings, from highly skewed splits to distributions closer to IID.


\end{document}